\documentclass[journal]{IEEEtai}
\usepackage{comment}
\usepackage[pagebackref]{hyperref}
\usepackage{amsmath}  
\usepackage{subfigure}
\usepackage{amssymb}
\usepackage{algorithm}
\usepackage[noend]{algorithmic}
\usepackage{wrapfig}
\usepackage{graphicx}
\usepackage{caption}
\usepackage{multirow}
\usepackage{color}
\usepackage{cite}
\usepackage[table]{xcolor}    

\renewcommand{\AA}{{\cal A}}
\renewcommand{\figurename}{Fig.}
\definecolor{darkgreen}{rgb}{0.0, 0.6, 0.0}
\definecolor{darkblue}{rgb}{0, 0, 1.0}

\begin{document}

\title{Enhancing Vision Language Models with Logic Reasoning for Situational Awareness}
\author{Pavana Pradeep Kumar$^1$, Krishna Kant$^1$ and Suya You$^2$\thanks{$^1$CIS Department, Temple University, Philadelphia, PA, USA. (e-mail:  pavana.pradeep@temple.edu, kkant@temple.edu)}\thanks{$^2$ARL, San Jose, CA. (e-mail:  Suya.you.civ@army.mil)}\\ 
\thanks{This research was supported by the ARL grant 272411} }
\IEEEoverridecommandlockouts
\IEEEpubid{\makebox[\columnwidth]{978-1-5386-5541-2/18/\$31.00~\copyright2018 IEEE \hfill} \hspace{\columnsep}\makebox[\columnwidth]{ }}
\maketitle
\IEEEpubidadjcol

\begin{abstract}
Vision-Language Models (VLMs) offer the ability to generate high-level, interpretable descriptions of complex activities from images and videos, making them valuable for situational awareness (SA) applications. In such settings, the focus is on identifying infrequent but significant events with high reliability and accuracy, while also extracting fine-grained details and assessing recognition quality. In this paper, we propose an approach that integrates VLMs with traditional computer vision methods through explicit logic reasoning to enhance SA in three key ways: (a) extracting fine-grained event details, (b) employing an intelligent fine-tuning (FT) strategy that achieves substantially higher accuracy than uninformed selection, and (c) generating justifications for VLM outputs during inference. We demonstrate that our intelligent FT mechanism improves the accuracy and provides a valuable means, during inferencing, to either confirm the validity of the VLM output or indicate why it may be questionable. 
\end{abstract}

\begin{IEEEImpStatement}
This paper applies emerging Vision Language Models (VLMs) for vision-based situational awareness in cyber-physical environments, focusing on safety, security, and policy compliance. While VLMs offer high-level descriptions of infrequent but critical events, traditional computer vision (TCV) methods better capture finer details like people, objects, locations, and movements. We explore integrating VLM and TCV through logical reasoning to enhance situational awareness, improve fine-tuning (FT) efficiency, and ensure output reliability. Fine-tuning VLMs can be costly, especially for infrequent events. We propose a technique that not only makes the FT efficient but also offers a way for sanity checking of the outputs during inferencing. The latter ensures enhanced reliability in recognizing crucial events, which is critical in the situational awareness context. 
\end{IEEEImpStatement}


\section{Introduction}
\label{s:intro}

Many cyber-physical systems routinely use Video-based monitoring of the premises with varying levels of automation in detecting events that require closer attention by humans. It is desirable to automate the process fully and provide a description of the situations that may be notable in some way. For example, in traffic monitoring on the road, it may be desirable to detect accidents, near-accidents, criminal activities using vehicles, etc. These situations are less common than normal traffic but of primary interest in monitoring. Although the notable or {\em target} events/activities are typically of negative variety (e.g., safety, security, policy violations), they can be arbitrary (e.g., monitoring who enters or exits a building or room). We assume throughout this paper that the target activities/events are explicitly described to enable suitable monitoring. We henceforth call the set of those as the {\em target task}.

Vision Language Models (VLMs), built on the foundation of Large Language Models (LLMs) have shown impressive abilities to recognize and describe complex scenes.  However, for good accuracy, VLMs (and LLMs) generally need to undergo task specific tuning either in form of instruction tuning  (i.e., showing it sample inputs and outputs) or fine-tuning (modification of neural net weights)~\cite{wu2025llm}. Here we focus on fine-tuning (FT) which is generally quite resource intensive. It is thus important to intelligently select inputs for situations where the VLM is performing poorly. In situational Awareness (SA) applications, many of the target activities are likely to be relatively infrequent (e.g., accident vs. normal traffic) and thus an informed selection of input video segments that are likely to contain the desired activities becomes important. If the selection can be done prior to labeling the selected segments, we not only save on processing but also on labeling effort. 

The key attraction of using VLMs for SA is that they can provide semantically rich descriptions of scenes that go well beyond what is reasonably achievable using Traditional Computer Vision (TCV) techniques. Here, TCV means deep learning networks such as CNNs or transformers specifically designed and trained for purposes like recognizing objects, object attributes, specific actions, etc.~\cite{manakitsa2024review}. The TCV methods still remain important and can be exploited in at least three ways: (a) For extraction of fine-grain, quantitative details of the situation that VLMs generally have difficulty with, (b) To intelligently pick the video segments for efficiently fine-tuning the VLMs for the target task, and (c) To help in producing a justification for the  recognized activities (when deemed to be reliable) or reasons why the output may not be reliable. All these uses require an integration of VLM and TCV outputs in a generic way, for which we exploit logical reasoning conducted efficiently to meet the real-time situational awareness objectives. 

To the best of our knowledge, this is the first work of its type to integrate explicit logic reasoning with VLM and TCV to make three substantial contributions in the context of VLM-based situational awareness: 
\begin{enumerate}
\item A mechanism to augment VLM output with fine-grain details.

\item An intelligent (or {\em directed}) FT mechanism that yields much better accuracy than an uninformed or random selection (which we henceforth refer to as {\em undirected} selection). This happens even when the length of the FT period is kept the same, and 

\item A mechanism to produce justification of VLM output during the inferencing phase (including an indication when the output is likely unreliable).
\end{enumerate}


The rest of this paper is organized as follows. Section~\ref{s:integration} presents the detailed design of our directed FT mechanism. Section~\ref{s:evaluation} discusses the experimental assessment of the directed FT against the undirected one. Section~\ref{s:related} discusses the related work. Finally, section~\ref{s:conc} concludes the discussion.

\section{Proposed Framework}
\label{s:integration}

\subsection{Motivation for VLM/TCV Integration}

The key aspect in situational awareness (SA) is quick detection and tracking of of not only the unusual events but also a host of other that are important to ensure safe operation of the monitored infrastructure. VLMs are excellent at providing an accurate high-level description of the situation, but their LLM based design is not intended for fine-grain quantitative information such as precise locations, distances, orientations, movements, etc.  Such details are often not only necessary in SA applications but can also be exploited in other ways including VLM fine-tuning and output justification, as detailed in this paper. Furthermore, so long as we limit ourselves to simple aspects such as objects and their postures/attributes, simple actions, and quantitative information (e.g., location, speed, count, etc.), these can all be handled easily by a combination of well-established, pretrained TCV algorithms plus simple vision and reasoning operations (e.g., perspective correction, location/distance estimation, frame-to-frame tracking of movements, etc.) Although some additional data labeling, training, or tweaks to existing TCV algorithms may be required, the effort involved is of similar level as tweaks to VLM pipeline and its fine-tuning. However, to capture both high-level and fine-grain features of a situation, we need to match up the outputs of VLM and TCV and integrate them together for which we make use of logic reasoning -- in particular, the satisfiability modulo theories (SMT). In fact, we also make use of an auxiliary VLM to further assist with efficient FT without labeling effort and deeper justification as described in subsequent sections.

\subsection{Tasks and Activities}

\begin{wrapfigure}[10]{R}{0.63\linewidth}
\includegraphics[width=\linewidth]{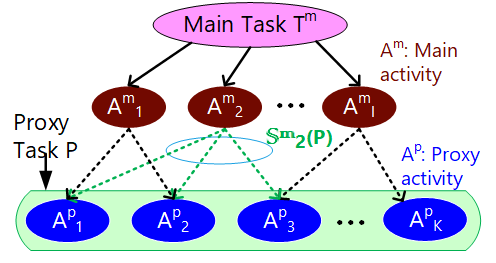}
\vspace{-18pt}
\captionof{figure}{Task and Activity Relationship\label{f:activities}}
\end{wrapfigure}

We start with a {\em predefined} target task, henceforth called as {\bf main} task, $\AA^m$, consisting of a set of ``activities" (or situations) that we want to recognize and track. \footnote{Although we only consider predefined activities, some of these may actually be learned automatically through VLM descriptions in the past.} We also define an {\em auxiliary task} $\AA^a$, which contains a variant of every activity in $\AA^m$. Accordingly, we have two VLMs, a {\em main VLM} (VLM$^m$) and an auxiliary VLM (VLM$^a$) to recognize them. We also define a {\em proxy task} $\AA^p$ that contains a set of {\em simple} activities that will provide a rather loose characterization of $\AA^m$ and $\AA^a$ as described below. Auxiliary activities are largely intended to be recognizable by standard, pretrained TCV algorithms -- although, tweaks to the algorithms and/or additional labeled training data may also be required in some cases. With this, our FT and justification approach revolves around consistency across the outputs for these three tasks. Note that while $\AA^m$ is given, both $\AA^a$ and $\AA^p$ can be constructed largely automatically. Table~\ref{t:task} shows the notations for the 3 tasks and the corresponding activities. Here $A^a_i, i=1..I$ are defined so that there is 1-to-1 correspondence between $A^m_i$ and $A^a_i$'s, and $i$ could thus be considered as a ``class" for recognition purposes. However, the proxy activities are interpreted differently as illustrated in Fig.~\ref{f:activities}.  It shows that each activity $ A^m_i$ of task $\AA^m$ is associated with some subset $\mathbb{S}^m_i(A^p)$ of proxy activities. For example, $A^m_2$ has three associated proxy activities $A^p_1..A^p_3$. Note that a similar depiction applies to the auxiliary task as well, except that the subsets $\mathbb{S}^a_i(A^p)$ would be different.

\begin{figure}[htb]
\small
\vspace{-6pt}
\begin{tabular}{|l|l|} \hline
{\bf Task type} & {\bf Defined via}\\ \hline
Main Task & Main activity set ($\AA^m=\{A^m_i,i=1..I$\})\\ \hline
Auxiliary Task & Auxiliary activity set ($\AA^a=\{A^a_i,i=1..I$\})\\ \hline
Proxy Task & Proxy activity set ($\AA^p=\{A^p_k,k=1..K$\})\\ \hline
\end{tabular}
\captionof{table}{\small Task and Activities\label{t:task}}
\end{figure}

We now place the requirement that $A^m_i\implies \mathbb{S}^m_i(A^p)$. Here, $\implies$ means that if the activity $A^m_i$ is observed, all of the (simpler) activities in the set  $\mathbb{S}^m_i(A^p)$ should also be observed. That is, the set $\mathbb{S}^m_i(A^p)$ is necessary for $A^m_i$ to occur but may not be sufficient. The exposition here is informal; in actual implementation, such requirements and the underlying activities are represented as {\em logic assertions} as discussed later. 

The intent is to use TCV to recognize $\mathbb{S}^m_i(A^p)$ and exploit the one-way implication as a weak but efficient consistency check with the VLM output. For TCV, we use standard algorithms such as YOLO variants to detect objects/poses, potentially augmented with additional neck/head layers in YOLO series of object detectors for recognizing additional attributes if needed~\cite{jegham2024yolo}. Any movement tracking can be done by frame-to-frame monitoring of objects, perspective transformations, and reasoning about the locations/movements of objects. In particular,  if a person/machine disappears from view temporarily and returns, we can track it easily and accurately in most cases. For simplicity, we also use the same set of proxy activities for the auxiliary VLM. That is, $A^a_j\implies \mathbb{S}^a_j(A^p)$, where $\mathbb{S}^a_j(A^p)$ is some non-null subset of $A^p$. 

Since we use TCV largely for simple and well-studied purposes, we expect quite high TCV accuracy; however, in case of a mistake by either or both TCV and VLM, it is extremely unlikely that they will still be consistent. Thus, mistakes only amount to some further fine-tuning that may not be necessary. Also note that in a continuous video monitoring environment, we can easily exploit temporal consistency to correct transient mistakes on the fly. For example, if over a few frames a truck is recognized as a car or not detected at all, and then we get it back in subsequent frames, the correction is straightforward and easily supported by our logic reasoning mechanism. However, it does require maintaining a buffer of recognition results from the last $K$ frames where $K$ is chosen as a tradeoff between output delay and temporal correction ability.

\subsection{Consistency Driven Fine-tuning}
\label{s:approach}

Given the setup above, we can process the video frames through VLM$^m$, VLM$^a$, and TCV, and determine, via spatio-temporal logic reasoning the following: (A)  Whether the VLM$^m$ output class, say $i$ is same as the  VLM$^a$ output class, say $i'$, (B) Whether all the activities in the set $\mathbb{S}^m_i(A^p)$ are observed and (C) Whether all the activities in the set $\mathbb{S}^a_{i'}(A^p)$ are also observed. If any of these {\em consistency conditions} do not hold, we conduct an intelligent or {\bf directed} selection of ``inputs" (or video segments) for further fine-tuning of the VLM$^m$ and VLM$^a$. 

The process can be repeated in a loop until the FT performance saturates or is terminated for other reasons such as running out of FT Data (FTD). We henceforth call the data used to evaluate FT needs as {\em evaluation data} (ED). Note that the conditions (A)-(C) do not need any labeled data. That is, {\em the determination of whether we need more fine-tuning or not can be done without first labeling the data.} Once we determine that more fine-tuning is needed, the FT itself needs labeled data, as usual. This allows for a more focused data labeling effort. Of course, if the ED were labeled, we could simply check if the predicted class $i$ is same as the label. In the following we show an interesting result that the purely {\em consistency-driven} evaluation approach provides almost the same quality of FT result as  {\em accuracy-driven} (i.e., one with labeled ED).

\subsection{Related Work}
\label{s:related}

To the best of our knowledge there is no work in the literature that exploits direct logic reasoning and/or TCV for enhancing fine-tuning and justifiability; however, there is plenty of work on individual topics.

Our work here does not concern itself precisely how the fine-tuning is done, and thus applies regardless of the underlying method. For example, parameter efficient fine-tuning (PEFT) attempts to minimize the parameters (weights) that are  altered~\cite{zhang2025parameter}. The typical methods to fine-tune include addition of a classifier on top of the visual backbone~\cite{radford2021learning}, an additional feature adapter~\cite{gao2024clip}, or using prompts along with VLMs' chain of Thought (CoT) reasoning adjust classification probabilities.  

The logical reasoning used in this paper involves the extension of first-order deductive reasoning using explicit Rules of Inference (RoIs). This differs from other recent notions of reasoning using VLM/LLM outputs~\cite{patil2025advancing, bandyopadhyay2025thinking, yang2023enhance}. Such reasoning fine-tunes the VLM/LLM to ask questions and generate answers directly, potentially in conjunction with external information obtained via web searches. Although the LLM reasoning abilities continue to improve, the very notion of reasoning by LLMs has been questioned~\cite{patil2025advancing}.

In the space of TCV and, more generally, deep learning, the issue of reasoning is often described as neuro-symbolic AI~\cite{bhuyan2024neuro, colelough2025neuro, lee2022neuro}. This form of AI attempts to enforce the constraints indirectly via the loss function of the neural net. For example, the popular Logic Tensor Networks (LTN)~\cite{badreddine2022logic} enforces logic constraints implicitly and approximately by using differentiable extensions of Boolean operations~\cite{van2022analyzing}) to avoid the problem of exploding or vanishing gradients. Explicit logic reasoning approaches are relatively sparsely explored~\cite{artikis2010logic, ramirez2013enhancing}. Ref~\cite{Pavana-safety, Pavana-driver} attempts to use explicit reasoning for accident and driver behavior characterization.

Explainable AI has seen a burgeoning amount of literature~\cite{hosain2024explainable}. 
Although much of it concerns explaining the AI's decisions, the focus has now expanded to the more important problem of justifiability of those decisions~\cite{holzinger2020xxai}. Our method not only supports justifiability in a simple way but also gives indications when the output may not be reliable.

\subsection{Justifying Inferences}

In many SA applications, the trustworthiness in the monitoring target activities is crucial. Thus, a {\em justification} of the result (when deemed reliable) or an indication of unreliability is essential. The proposed {\em consistency-driven} FT mechanism can be adapted for this role by simply continuing the consistency checks during the inference phase. In this case, we could run the two VLMs in parallel on two different GPUs, if available, or forego VLM$^a$ during inferencing and thus weaken the validation. A third choice is to use VLM$^a$ only for ``difficult scenarios". Such scenarios can be learned from the history of observed inconsistencies, but we do not pursue that aspect here. We later quantify the overhead of justification, which should be possible to do in real time with improvements in hardware speeds.

\section{Implementing Directed Fine Tuning}

\subsection{Identifying Proxy and Auxiliary Activities}
\label{s:proxy}

Since the main activities of interest could be arbitrary, they are best assumed to be given (though some of them can be learned). However, it is convenient to generate the corresponding proxy and auxiliary activities automatically. The proxy activities can be seen as the breakdown of a main activity in terms of simpler lower-level activities that they are composed of, whereas auxiliary activities can be considered as an alternate description of the main activity or that of very similar activity. Since these identifications need to be done only once at design time, the easiest method is to ask a highly capable multimodal LLM  for both hierarchical decomposition and similar activity identification. The LLM can also be exploited to discard descriptions that are flawed. For example, Video-chatGPT~\cite{Video-ChatGPT2024} can be used not only for generating alternate description but also to check semantic alignment between the main and auxiliary descriptions. Video-chatGPT produces scores (using a scale of 0-5) in 4 different areas: correctness, consistency, orientation, contextual understanding, and understanding of temporal changes. 

In most cases, the hierarchical decomposition ultimately concerns main objects in the scene and their properties (poses, attributes, movement, spatio-temporal relationship to another object, etc.) For example, the rather complex activity of driving a vehicle can be characterized as person, vehicle, and their relative orientation/location/movement. As discussed earlier, the characterization is a necessary but not sufficient condition for the driving activity. 

We next automatically determine the set of proxy activities implied by the VLM recognized activity $A_i$, denoted by the mapping $\mathbb{S}_i (A^p)$. (We have omitted the superscript $m$ or $a$ here since the same procedure applies to both.) First, we express every proxy activity $A^p$ as a logic assertion, say $\psi_{A^p}$. Let $D$ denote the entire input dataset available for fine-tuning, which is a labeled set of short video segments. Then, for each class $i$, we pass each input through TCV and identify which assertions in $\psi_{A^p}$ hold for it. In the last step, for each class $i$, we pick the assertions that hold in most (e.g., 90\%) cases. This set then characterizes the set $\mathbb{S}_i$ for each $i$. Some fuzziness in considering the assertions (i.e., the 90\% part) is necessary in real-world settings. A different way to allow fuzziness (which we have not done) is by attaching a strength measure with each assertion which indicates a belief in the assertion, perhaps quantified by a suitable similarity metric.

\subsection{Using Explicit Logic Reasoning}
\label{s:bg-logic}

Explicit logic reasoning has been used successfully in numerous domains~\cite{abraham2016satisfiability} and contexts~\cite{Pavana-safety, Pavana-IoT, Pavana-driver, Pavana-posture}. The most basic use is first-order logic extended with additional ``theories" to reason about relevant topics such as integer/real arithmetic, laws of linear/angular motion, spatial relationships, etc. The reasoning can be done by highly popular Satisfiability Modulo Theory (SMT) based tools such as Z3~\cite{Z3-SMT-2008} and YICES~\cite{dutertre2014yices}, which can routinely solve very large practical problems. The three crucial parts in an SMT model are (a) Rules of Inference (RoIs), (b) Various real-world constraints (e.g., walking speed $<2$ m/s), and (c) Groundings or facts determined from the environment (e.g., by analysis of objects in individual frames along with perspective correction). We can also define higher-level concepts as reusable functions suitable in the ``logic program". For example, consider the function ``following(V1, V2)'' that asserts that vehicle V1 is following vehicle V2. This can be decomposed, if needed, as the relative separation between V1  and V2 in a sequence of frames. It can also be defined as an uninterpreted function whose truth value is determined via TCV, or even VLM. Note that explicit logic reasoning discussed here is different from other forms of reasoning as discussed under related work (section~\ref{s:related}), and neither involve loss functions, nor any training. 

Although simple SMT-based reasoning is adequate for this paper, it can be easily extended to cover temporal reasoning (e.g., a situation holding until ended by some event) or real-time relationships like timeouts. This adaptability is showcased through several temporal extensions such as~\cite{konur2013survey,  artikis2014event, vlassopoulos2017towards} as used in~\cite{Pavana-safety, Pavana-driver}. In particular, the reasoning can include both  ``hard" clauses (i.e., must-hold) and ``soft" clauses (nice to have based on a weight), as explored in~\cite{Pavana-safety}.

\subsection{Fine Tuning Algorithm}
\label{s:integration-FT}

\begin{figure}[htb]
 \vspace{-6pt}
 \begin{center}
    \includegraphics[width=\linewidth]{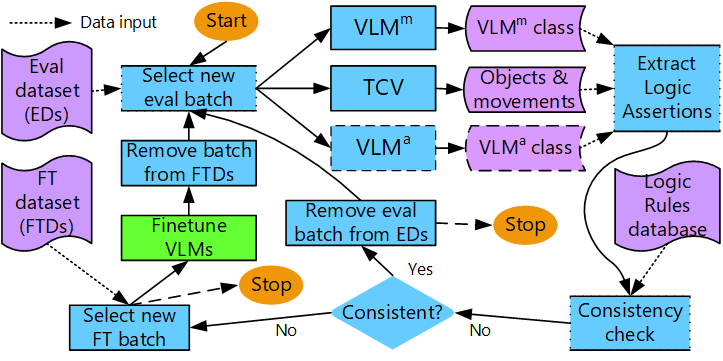}
  \end{center}
\vspace*{-12pt}
\caption{Application Directed Fine-tuning of VLMs} \label{f:FT-illus}
\vspace{-6pt}
\end{figure}

With all the pieces in place, we describe the overall FT algorithm as illustrated in Fig.~\ref{f:FT-illus}. The flowchart and description here are somewhat simplified for clarity, and variations may be used in an actual implementation. The purple boxes show the input/output data and the blue/green boxes show operations. We start with two datasets: the FT data (FTD) and the evaluation data (ED). As stated earlier, ED need not be labeled,  although it is labeled in our case to allow the study of both consistency-based and accuracy-based reasoning introduced above. Ideally, the ED should include one or more batches for each VLM$^m$ class, whereas each FTD batch focuses on a specific class(es) of VLM$^m$ that fail the consistency checks. {\em Note that the ED is used during fine-tuning, and thus differs from the test dataset used for eventual performance evaluation.}

The shown ``Start" of the FT loop uses a batch of inputs from ED, depicted as {\em eval-batch} in Fig.~\ref{f:FT-illus}, for each evaluation. It runs the eval-batch through VLM$^m$, VLM$^a$ (if used), and TCV inferencing steps and collects the results. The outputs provide the ``groundings" for the relevant assertions in the logic representation of the detected classes and proxy activities and enable consistency checks with the help of the Logic rules database, which is prebuilt. Next, we use SMT for consistency check as shown. If no inconsistency is observed for an eval-batch, we remove it from ED since it is not interesting for finding inconsistencies.
In case of inconsistency, the SMT framework provides the offending assertions (i.e., the subset of assertions that failed) which can be mapped to the VLM class(es) for which more fine-tuning is needed. Thus, the next step is to choose a batch from FTD, depicted as {\em FT-batch} in Fig.~\ref{f:FT-illus}.

As an illustration consider a video frame in Fig.~\ref{f:time-all} from TU\_DAT dataset showing a rear-end accident. The TCV (YoLovx) gives the bounding boxes of all the cars, and we can assign them some pseudo-IDs, such as car1, car2, etc. The recognized VLM$^m$ class here would be ``1" as defined by our main task activities in the left column of Table~\ref{t:accident-classes}). The VLM$^a$ class will also be ``1" (right column in Table~\ref{t:accident-classes}). The  proxy activity set $\mathbb{S}^m_1(A^p)$ includes the activities (\#1,\#3) stated abstractly in Table~\ref{t:proxy-list}. These would be grounded in the analysis of the shown frame and a few prior frames in the video. Thus, the grounding of \#1 (``A car behind another car in the same lane") will require the pseudo-IDs of the two vehicles and the truth value of the statement. Thus, after grounding, we obtain a statement like: ``car2 moving behind car1 in the same lane". This grounding will be done in the logic domain (not natural language), and it will mark the corresponding assertion as true. Thus SMT can be used to check for various consistency conditions easily. 

To avoid clutter, Fig.~\ref{f:FT-illus} only marks transition to ``stop" in two places to indicate termination, which would happen if we run out of FTD or ED data, exceed a time limit or see little further improvement in accuracy. We start our directed FT after some initial FT with a small set of labeled inputs. For fine-tuning we chop longer videos into small ones so that each video focuses on only one class of interactions as far as possible.\footnote{Even with very short video segments, it is possible to have more than one activity. Such situations can be recognized by defining additional composite classes for conditions that are likely to occur together sufficiently often. It is expected that the number of such combinations will be small.} We label (or caption) these video segments according to the requirements of the specific VLM used. For image-based VLMs, we label each frame in the video segment identically.

\section{Details of Experimental Setup}
\label{s:evaluation}

\subsection{VLMs Used For Evaluation}
\label{s:VLMS-considered}

Although our primary interest is in video-based VLMs, we have also evaluated our finetuning methodology on image-based VLMs~\cite{zhang2024vision}. We also consider both LLM-backed VLMs and those that merely do image-text matching. Each of these uses a different strategy to align vision and language models. By analyzing these diverse models, we aim to derive conclusions with a good degree of generality.

MiniGPT-4~\cite{zhu2023minigpt4} an image based VLM defines two trainable layers for aligning a frozen vision transformer model with a frozen LLM model. MiniGPT4-Video~\cite{ataallah2024minigpt4} is an extension of MiniGPT-4 that processes both videos and textual conversations. X-CLIP~\cite{ma2022x} is a popular video-text matching-based VLM for video-text retrieval and generates multi-grained visual and textual representations. Video-LLaMA~\cite{zhang2023video} is a state-of-the-art multi-modal framework that allows LLMs to comprehend audio and video signals. Numerous other VLMs exist and continue to emerge rapidly but we believe that our chosen set represents the range quite well. 

In recent years, the selective state space model (SSM)~\cite{wang2024state} has emerged as an appealing alternative to transformer-based VLMs. For example, VideoMamba~\cite{li2024videomamba} is a purely SSM-based model tailored for video understanding, and integrates the advantages of convolution and attention. It provides a linear-complexity method, which is attractive from a real-time situational understanding perspective.

\subsection{Description of Datasets Used}
\label{s:datasets}

\begin{figure}[htb]
\centering
\begin{minipage}{0.5\linewidth}
\includegraphics[width=\linewidth]{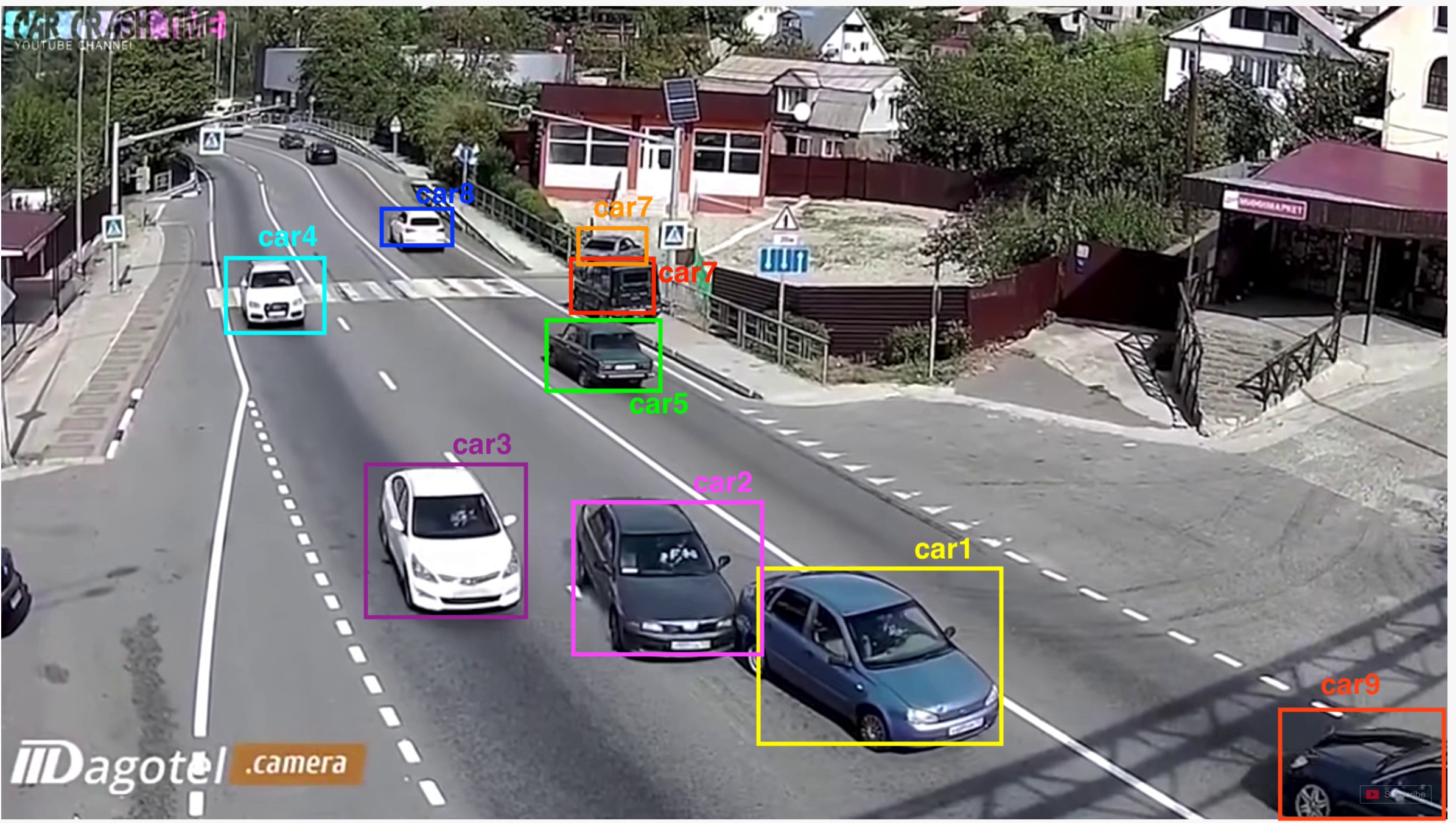}
\caption{Finding key objects in the scene}
\label{f:time-all}
\end{minipage}~
\begin{minipage}{0.44\linewidth}
\vspace{-6pt}
\includegraphics[width=0.5\linewidth]{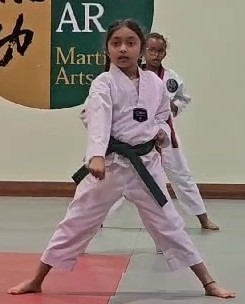}~
\includegraphics[width=0.5\linewidth]{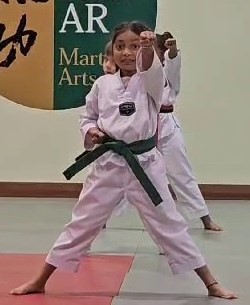}
\vspace{-12pt}
\caption{Green belt movements in Taekwondo }
\label{f:taekwondo-data}
\end{minipage}
\vspace{-18pt}
\end{figure}

We evaluated our FT methodology using three very different datasets. The first two involve limited activities, whereas the third one is much more open-ended. Our first dataset, called TU\_DAT~\cite{TU_Dat}, concerns road traffic and contains videos collected in challenging driving and weather conditions. Fig.~\ref{f:time-all} (included later in the evaluation section)  shows a scene from this dataset indicating a rear-end accident. 

\begin{figure}[htb]
\center
\subfigure[]
{\includegraphics[height=0.95in,angle=0]{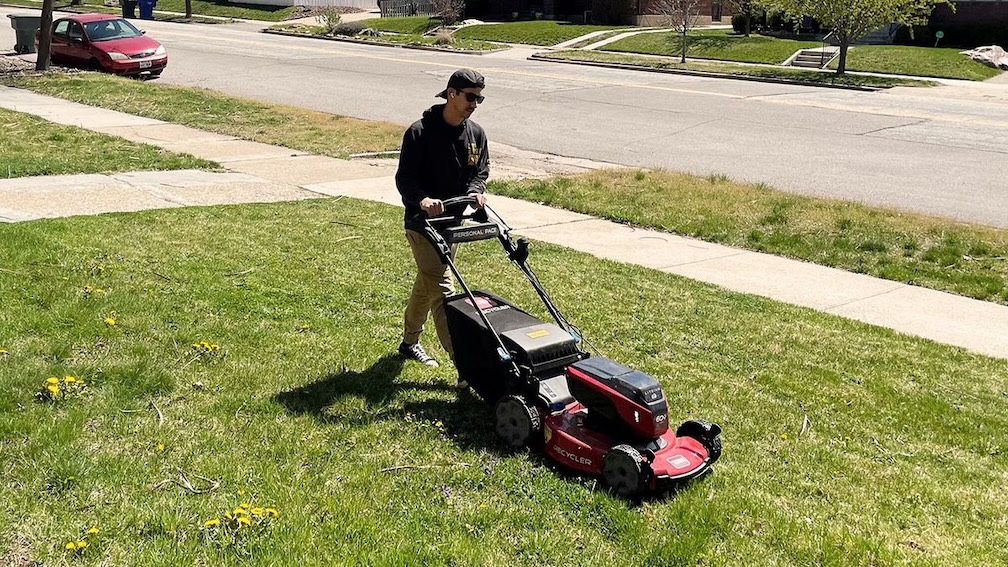}}~
\subfigure[]
{\includegraphics[height=0.95in,angle=0]{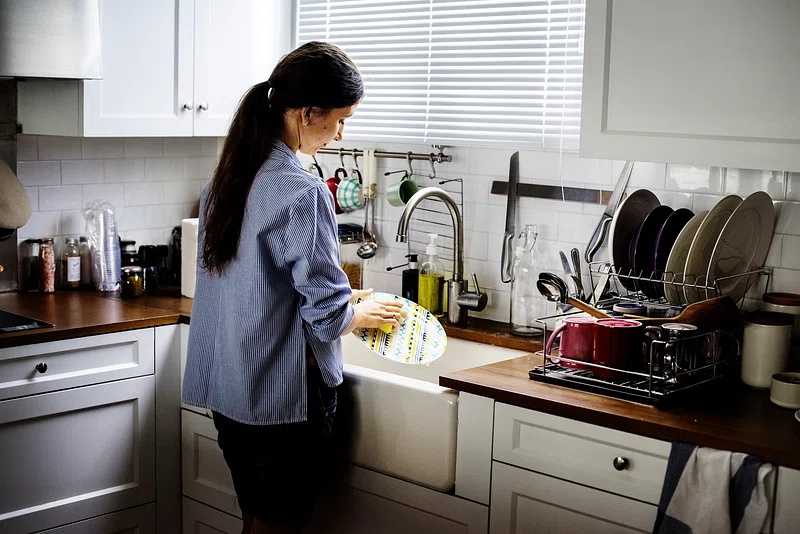}}
\vspace{-6pt}
\caption{Few activities from Kinetics dataset
\label{f:kinetics-data}}
\vspace{-12pt}
\end{figure} 

Our second dataset, the Taekwondo dataset, captures movements performed by Taekwondo athletes. Understanding the movement patterns is a crucial component of Taekwondo training, as explained in the Taekwondo America student manual~\cite{Taekwondo}. This dataset has 35 videos, which feature either a single student or multiple students performing the movements in sequence for each belt pattern. Fig.~\ref{f:taekwondo-data} (a) shows the walking stance, low block, and (b) shows the walking stance, reverse punch of a student in a dark green belt pattern. 

\begin{wraptable}[6]{R}{0.45\linewidth}
\vspace{-12pt}
\footnotesize
\caption{Dataset details}\label{t:details-datasets}
\vspace{-6pt}
\begin{tabular}{|c|c|c|c|} \hline
{\bf Dataset} & {\bf FT } & {\bf Eval} & {\bf Test} \\
{\bf type} & {\bf Data} & {\bf Data} & {\bf Data}  \\ \hline
TU\_DAT & 200 & 75 & 45\\ \hline
Taekwondo & 105 & 65 & 30\\ \hline
Kinetics & 2000 & 800 & 350\\ \hline
\end{tabular}
\end{wraptable}

Our third dataset is the Kinetics-100~\cite{kay2017kinetics}, a large video dataset focused on human actions. The list of action classes includes single-person actions (e.g., drawing, drinking), person-person actions (e.g., hugging, shaking hands), and person-object actions (e.g.,  opening gifts, mowing lawn, washing dishes). Kinetics-100 has 100 human action classes, with 400–1150 clips for each action, and each clip lasts around 10 seconds. Fig.~\ref{f:kinetics-data} shows some activities from the Kinetics-100 dataset, such as (a) Lawn mowing and (b) Washing dishes. 

We use several augmentation methods to acquire enough data volumes for both TU\_DAT and Taekwondo datasets. We do this by using Keras that provides various methods for real-time image augmentation. The employed augmentations encompass flipping, translation, shear, and rotation. Table~\ref{t:details-datasets} shows the details of all three datasets.

\subsection{Classes Used to Fine-Tune VLMs} 

\begin{table*}[htb]
\footnotesize
\begin{minipage}{5.2in}
\begin{tabular}{|p{0.10in}|p{1.4in}|p{3.2in}|}
\hline
No. & \multicolumn{1}{c|}{Classes in VLM$^m$} & \multicolumn{1}{c|}{Classes in VLM$^a$}          \\ \hline
1  & Car hit by another from behind & Car moving in same direction and one behind another             \\ \hline
2  & Car hit by another from side   & Car moving in opposite direction and perpendicular to another            \\ \hline
3  & Car hit by another from front  & Car moving in opposite direction      \\ \hline
4  & Car hits a static object             & Car moving very close to static object \\ \hline
5  & Motorcycle hits a pedestrian         & Motorcycle moving very close to a walking pedestrian    \\ \hline
6  & Traffic videos                       & Cars moving next to, across, or behind cars in same or opposite direction  \\ \hline
7  &  Not defined                         & Car \& motorcycle moving next to one another \\ \hline
8  &  Not defined                         & Pedestrians walking                            \\ \hline
\end{tabular}
\end{minipage}
\begin{minipage}{1.8in}
\begin{tabular}{|p{1.4in}|r|}
\multicolumn{1}{l}{{\bf Score Range (low - high)}\vspace{3pt}} & \multicolumn{1}{l}{{\bf 0-5}\vspace{3pt}} \\ \hline
\textbf{Average measure of performance} & \multicolumn{1}{c|}{Score}         \\ \hline
Correctness score  &   4.09  \\ \hline
Consistency score  & 4.02\\ \hline
Detailed orientation score  &  3.91   \\ \hline
Contextual understanding score  & 3.97 \\ \hline
Temporal understanding score  & 4.01\\ \hline
\end{tabular}
\end{minipage}
\caption{(a) Class description for TU\_DAT, (b) Similarity Scores for VLM$^m$ and VLM$^a$ classes}
\label{t:accident-classes}
\end{table*}

\begin{table}[htb]
\footnotesize
\caption{Classes Used for Taekwondo Dataset}
\label{t:taek-classes}
\begin{tabular}{|p{0.10in}|p{1.55in}|p{1.5in}|}
\hline
No. & \multicolumn{1}{c|}{Classes in VLM$^m$}                 & \multicolumn{1}{c|}{Classes in VLM$^a$}   \\ \hline
1  & Left leg still, right leg still        & Left arms and right arms out        \\ \hline
2  & Left leg still, right leg forwards     & Left arms out, right arms folded     \\ \hline
3  & Left leg still, Right leg backwards    & Left arms folded, right arms out     \\ \hline
4  & Right leg still, Left leg forwards     & Left arms and right arms folded  \\ \hline
5  & Right leg still, Left leg backwards   & Left arms on the head, right arms folded \\ \hline
6  & Left leg forward, Right leg backwards & Right arms on the head, left arms folded \\ \hline
7  & Right leg forward, Left leg backwards & Not defined  \\ \hline
\end{tabular}
\end{table}

\begin{table}[htb]
\footnotesize
\caption{Description of Classes Used for Kinetics Dataset}
\label{t:kinetics-classes}
\begin{tabular}{|p{0.10in}|p{0.45in}|p{2.6in}|}
\hline
No. & VLM$^m$ Classes & VLM$^a$ Classes   \\ \hline
1  & Arm wrestling &  Using one arm in gripping the opponent's hand, applying pressure, stabilizing the body with both the legs, adjusting posture for leverage   \\ \hline
2  & Baking cookies & Mixing ingredients, rolling out dough, shaping cookies with hands, using legs to reach the oven, and opening it to place the baking tray inside \\ \hline
3  & Brushing teeth & Using one hand to hold the toothbrush, apply toothpaste with other hand, and move it in circular motions to clean the teeth and gums\\ \hline
4 & Cart wheeling & Both hands push off the ground and support the body while the legs kick up and rotate in the air to complete the cartwheel\\ \hline
5 & Cheer leading & Using both hands to perform motions like clapping, waving, or holding pom-poms, while legs execute jumps, kicks, and stunts to enhance cheer routines\\ \hline
6   & Moving Lawn &  Using one of the hands to start the mower, adjust steering and cutting height, using both legs in pushing or guiding the mower and walk around obstacles   \\ \hline
7   & Washing dishes & Using both hands to scrape the dishes, applying soap to the dishes and rinsing the dishes with water       \\ \hline
\end{tabular}
\end{table}

The TU\_DAT dataset contains several accident scenarios in road traffic, forming the classes for fine-tuning a VLM. Since our proposed method includes fine-tuning two VLMs, the videos in the TU\_DAT dataset have been categorized into modeling accident scenarios for VLM$^m$ and recognizing the relative position/movements of vehicles for VLM$^a$. The description of classes used in fine-tuning VLM$^m$ and VLM$^a$ on TU\_DAT are shown in Table~\ref{t:accident-classes}(a). The table shows classes with the same number side by side for VLM$^m$ and VLM$^a$, reflecting a one-to-one relationship between them. As discussed earlier, VLM$^a$ descriptions were obtained using ChatGPT from VLM$^m$ descriptions (but shortened slightly for display in the table). 

To show how well the VLM$^m$ and VLM$^a$ descriptions are aligned we obtained the Video-chatGPT similarity scores for them, which we list in Table~\ref{t:accident-classes}(b). With all of the scores close to 4.0 (out of 5.0), it is clear that the two are closely aligned in all respects. We found a similar match for the other two datasets discussed below, but for brevity we do not list the similarity scores here. 

For the Taekwondo dataset, VLM$^m$ is fine-tuned to recognize the leg movements of the students, while VLM$^a$ is fine-tuned to identify the students' arm movements. The description of classes used in fine-tuning VLM$^m$ and VLM$^a$ on the Taekwondo dataset are shown in Table~\ref{t:taek-classes}. A VLM that understands Taekwondo rules can automatically generate arm movements corresponding to the leg movements.

For the kinetics dataset, we carefully chosen approximately 50 activities as ``target activities" out of the total of 100 activities. The reason to select so many target activities is to enable stress testing. As stated earlier, in reality, the target activities of interest are typically those that are unusual or anomalous in some way.  We omit a comprehensive description of all these 50 activities due to space constraints; instead, we present the descriptions of a selected few classes in Table~\ref{t:kinetics-classes}.

\subsection{Identification of Proxy Activities}

We illustrate the derivation of proxy activities here using TU-DAT; others are omitted due to lack of space. We start with a sample enumeration of simple activities, which in the context of road traffic are the spatial relationships between the key objects. These are listed informally Table~\ref{t:proxy-list}(a) and can be easily expressed as logic assertions. Note that we have expressed the relationships generically since we want to be able to check them for any pair of cars or a car and a pedestrian. The object IDs will come from processing actual videos, as described next. 

\begin{table}[htb]
\begin{footnotesize}
\caption{Illustrating proxy activity identification}\label{t:proxy-list}

\begin{minipage}{0.5\linewidth}
{\bf (a) Possible spatial relationships (informal)}\\

\begin{tabular}{|p{1.6in}|} \hline
A car behind another car in the same lane \\
A car facing another car in the same lane \\
A car moving closer to next car \\
A car moving into next lane in same direction \\
A car moving into next lane in opposite direction \\
A pedestrian in a traffic lane. \\
A car moving closer to a pedestrian \\ \hline 
\end{tabular}
\end{minipage}~~
\begin{minipage}{0.5\linewidth}
{\bf (b) Grounding of spatial relationships (informal)}\\ 

\begin{tabular}{|p{1.6in}|} \hline
car1 and car2 moving one behind another \\
car1, car2, car3 \& car4 traveling in same direction \\
car1 is following car2 at very close distance\\
car1 and car9 traveling in the opposite lanes \\
car7 is parked and not moving \\
car5, car8, \& car9 traveling in the same direction \\ \hline
\end{tabular}
\end{minipage}
\end{footnotesize}
\end{table}

\begin{table}[htb]
\begin{footnotesize}
\caption{Assertions for Reasoning}\label{t:assertions}
\begin{tabular}{|p{3.0in}|} \hline
\textbf{Variables:} car1, car2.., car9 are integers\\
\vspace{3pt}
\textbf{Functions:} Boolean, each with one Integer argument\\
move\_behind(), move\_very\_close(), move\_opp\_dirn() \\
move\_same\_dirn(), car\_moving(), car\_hit\_from\_behind()\\
\vspace{3pt}
\textbf{Groundings:}\\
move\_behind(car1,car2) $\wedge$ 
move\_very\_close(car1,car2) $\wedge$ \\
move\_same\_dirn(car1, car2) $\wedge$ move\_same\_dirn(car2, car3) $\wedge$ \\
move\_same\_dirn(car3, car4) $\wedge$ 
move\_same\_dirn(car5, car8) $\wedge$ \\
move\_same\_dirn(car8, car9) $\wedge$ 
move\_opp\_dirn(car1,car9) $\wedge$ \\
$\neg$ car\_moving(car7)  \\
\vspace{3pt}
\textbf{Proxy Assertion Example:}\\
car\_hit\_from\_behind (car1,car2) $\implies$ move\_behind(car1,car2) $\wedge$ move\_very\_close(car1,car2)\\
\hline
\end{tabular}
\end{footnotesize}
\vspace{-6pt}
\end{table}

We used the latest version of YOLO at the time to identify the key objects in the input images and track the objects across frames (in the case of videos) to ``ground" the situation in terms of objects, pseudo-object IDs, positions, distances, and relative movements. For example, for the scene in Fig.~\ref{f:time-all}, the identified (grounded) relationships are listed Table~\ref{t:proxy-list}(b) informally. Again, for actual processing, we need to turn these into proper logic assertions, as illustrated in Table~\ref{t:assertions}. The table shows the basic functions and their groundings for Fig.~\ref{f:time-all}. 

Given the groundings, we can determine which assertions in Table~\ref{t:proxy-list}(a) hold and for which cars. This in turn can be used to identify the assertions that are almost always true about the situation indicated by the video label, thereby giving us the required proxy assertions. For example, for a car to be hit from the back, the true assertions would be that the cars are moving one behind another and that they are are very close. This is also shown as the one-way implication in Table~\ref{t:assertions}.

\subsection{Fine-Tuning Specifics}

We start by illustrating the mechanics of our FT mechanism with an example shown in Table~\ref{t:FT-illus}. We use a rear-end accident scenario from the TU\_DAT dataset using XCLIP VLM. After the initial FT stage, VLM$^m$ and VLM$^a$ yield the captions as shown in the top two lines in Table~\ref{t:FT-illus}. The two outputs are inconsistent, which is affirmed by not having the expected class correspondence between VLM$^m$ and VLM$^a$. Therefore, we select and retrieve the videos with the labels as depicted for additional fine-tuning of both VLMs. 

\begin{table}[htb]
\begin{footnotesize}
 \caption{Illustrating Consistency Check in Fine Tuning\label{t:FT-illus}}
\begin{tabular}{|p{3.15in}|}\hline
{\bf VLM$^m$ Output:} Matching Caption: car hit by another from behind \\
{\bf VLM$^a$ Output:} Matching Caption: car moving in opposite direction\\     
\vspace{1pt}
{\bf Consistency Check:} NO - VLM$^m$ output is inconsistent with VLM$^a$ output \\
{\bf Retrieve Videos with labels:} car hit by another from behind, car moving in same directions and one behind another \\
{\bf Fine Tune VLM$^m$ and VLM$^a$} \\
{\bf Select a new eval batch:} To determine consistency between VLM$^m$ and VLM$^a$ outputs \\
{\bf Consistency Check:} YES - VLM$^m$ output is consistent with VLM$^a$ output \\
\hline
 \end{tabular}
\end{footnotesize}
\end{table}

For our results, we chose a fixed FT batch size of 20 videos. This is somewhat arbitrary, and one could vary the number as well. To make a fair comparison, we execute the loop four times for both directed and undirected cases, each using 20 videos. Again, undirected here means that the videos are chosen randomly without regard to the areas of FT weakness. We stop at four iterations since the improvement in consistency appears to be stalled after that. All experiments were performed on a workstation with two 48GB NVIDIA RTX A6000 GPUs. (Only one GPU was used in inferencing). Our results {\em do not} assume parallel inferencing  (see section~\ref{s:integration-FT}).

\section{Experimental Results}

We characterize the performance of the proposed FT framework in terms of 3 aspects: {\em Accuracy}, {\em Consistency}, and {\em Overhead} for the proposed framework as follows:
\begin{enumerate}
\item Accuracy: Fraction of test cases whose classification matches the ground truth. 

\item {\em Consistency Improvement Factor} (CIF): defined as $(n_b-n_e)/n_b$ where $n_b$ is the number of inconsistencies recorded before the FT, and $n_e$ is the number of inconsistencies at the conclusion of the FT procedure. 

\item Overhead: We measure the additional time due to the directed selection of inputs during FT and the justification time during inferencing. Ideally, this should be reported as a ratio of baseline time but we will report absolute times as well. 
\end{enumerate}

To ensure fairness, for the datasets with a small number of target activities (TU\_DAT and Taekwondo), we use the same number of FT iterations for both directed and undirected cases. For the Kinetics dataset, which has many (actually 50) target activities, we perform both directed and undirected fine-tuning for the same duration. 

The results in the following show that the directed method consistently outperforms the undirected one in all cases. Furthermore,  this differential applies to both image-based VLMs, such as Minigpt4~\cite{zhu2023minigpt} and its video-based version, MiniGPT4-video~\cite{ataallah2024minigpt4}. We have also demonstrated our FT mechanism on transformer-based VLMs that includes an LLM as the backend like Video-Llama~\cite{zhang2023video}, and more recent non-transformer, state space model(SSM) based VLMs like VideoMamba~\cite{li2024videomamba}. The same applies to models not backed by LLMs, such as XClip~\cite{ma2022x} and Video-MAE~\cite{wang2023videomae}. We also show that the improvement is sustained for two entirely different datasets, one concerning road-traffic and the other concerning Taekwondo classroom.

\subsection{Achieved Classification Accuracy}

The most important result for our (consistency-driven) FT approach is the classification accuracy achieved on the test-dataset. Table~\ref{t:acc-table} shows this for all three datasets and for both VLM$^m$ and VLM$^a$. It is seen that the directed FT significantly outperforms the undirected FT in {\em all} cases and for both VLMs. However, the results depend on the VLM and the dataset.

\begin{table}[htb]
\begin{footnotesize}
\caption{Accuracy of Consistency-Driven FT}
\label{t:acc-table}
\begin{tabular}{|c|c|cc|cc|}
\hline
\textbf{VLMs} &
  \textbf{Datasets} &
  \multicolumn{2}{c|}{\textbf{\begin{tabular}[c]{@{}c@{}}Undirected\\ \hline VLM$^m$ VLM$^a$\end{tabular}}} &
  \multicolumn{2}{c|}{\textbf{\begin{tabular}[c]{@{}c@{}}Directed\\ \hline VLM$^m$ VLM$^a$\end{tabular}}} \\ \hline
\multirow{3}{*}{MiniGPT4}    & TU\_DAT   & \multicolumn{1}{c|}{73.14} & 72.5  & \multicolumn{1}{c|}{82.14} & 82.5  \\ \cline{2-6} 
                             & Taekwondo & \multicolumn{1}{c|}{72.1}  & 71.7  & \multicolumn{1}{c|}{81.1}  & 81.45 \\ \cline{2-6} 
                             & Kinetics  & \multicolumn{1}{c|}{79.6}  & 78.4  & \multicolumn{1}{c|}{84.15} & 84.78 \\ \hline\hline
\multirow{3}{*}{MiniGPT4-V}  & TU\_DAT   & \multicolumn{1}{c|}{75.4}  & 74.15 & \multicolumn{1}{c|}{83.3}  & 83.55 \\ \cline{2-6} 
                             & Taekwondo & \multicolumn{1}{c|}{73.41} & 73.75 & \multicolumn{1}{c|}{82.81} & 82.6  \\ \cline{2-6} 
                             & Kinetics  & \multicolumn{1}{c|}{80.15} & 79.85 & \multicolumn{1}{c|}{84.4}  & 84.2  \\ \hline\hline
\multirow{3}{*}{Video-LLaMa} & TU\_DAT   & \multicolumn{1}{c|}{78.25} & 78.35 & \multicolumn{1}{c|}{85.12} & 85.84 \\ \cline{2-6} 
                             & Taekwondo & \multicolumn{1}{c|}{77.5}  & 77.85 & \multicolumn{1}{c|}{85.22} & 85.1  \\ \cline{2-6} 
                             & Kinetics  & \multicolumn{1}{c|}{82.15} & 82.85 & \multicolumn{1}{c|}{88.54} & 88.25 \\ \hline\hline
\multirow{3}{*}{Video-Mamba} & TU\_DAT   & \multicolumn{1}{c|}{76.6}  & 75.55 & \multicolumn{1}{c|}{81.3}  & 81.42 \\ \cline{2-6} 
                             & Taekwondo & \multicolumn{1}{c|}{76.41} & 76.8  & \multicolumn{1}{c|}{80.85} & 80.8  \\ \cline{2-6} 
                             & Kinetics  & \multicolumn{1}{c|}{80.15} & 79.35 & \multicolumn{1}{c|}{83.85} & 83.14 \\ \hline
\end{tabular}
\end{footnotesize}
\end{table}

Next, we compare our consistency-driven FT approach against the accuracy-driven approach on the test data. Recall that the accuracy-driven FT selects inputs based on the observed inaccuracies on ED (or evaluation data). The accuracy-driven FT does not require any TCV or VLM$^a$, but on the downside, it needs labeled ED and cannot provide justifiability during inference time. Table~\ref{t:acc-new-table} shows the accuracy-driven results for all three datasets. It is seen that the accuracy here is almost the same as in Table~\ref{t:acc-table}; {\em; thus, we are not losing anything by following the consistency-driven approach.}\footnote{The undirected results are slightly different than in Table~\ref{t:acc-table}, possibly due to randomness; they should ideally be the same.} 

\begin{table}[htb]
\begin{footnotesize}
\caption{Accuracy of Accuracy-Driven FT}
\label{t:acc-new-table}
\begin{tabular}{|c|c|cc|cc|}
\hline
\textbf{VLMs} &
  \textbf{Datasets} &
  \multicolumn{2}{c|}{\textbf{\begin{tabular}[c]{@{}c@{}}Undirected\\ \hline VLM$^m$ VLM$^a$\end{tabular}}} &
  \multicolumn{2}{c|}{\textbf{\begin{tabular}[c]{@{}c@{}}Directed\\ \hline VLM$^m$ VLM$^a$\end{tabular}}} \\ \hline
\multirow{3}{*}{MiniGPT4-V}  & TU\_DAT   & \multicolumn{1}{c|}{74.5}  & 75.5 & \multicolumn{1}{c|}{82.0}  & 82.52 \\ \cline{2-6} 
                             & Taekwondo & \multicolumn{1}{c|}{72.8} & 72.75 & \multicolumn{1}{c|}{82.1} & 81.5  \\ \cline{2-6} 
                             & Kinetics   & \multicolumn{1}{c|}{81.25}  & 79.5 & \multicolumn{1}{c|}{85.0}  & 85.25 \\ \cline{2-6} 
                             \hline\hline
\multirow{3}{*}{Video-Mamba} & TU\_DAT   & \multicolumn{1}{c|}{75.8}  & 76.25 & \multicolumn{1}{c|}{80.25}  & 80.0 \\ \cline{2-6} 
                             & Taekwondo & \multicolumn{1}{c|}{77.15} & 75.5  & \multicolumn{1}{c|}{80.12} & 80.35  \\ \cline{2-6} 
                             & Kinetics   & \multicolumn{1}{c|}{79.4}  & 80.55 & \multicolumn{1}{c|}{84.6}  & 84.1 \\ \cline{2-6} 
                             \hline 
\end{tabular}
\end{footnotesize}
\vspace{-18pt}
\end{table}

\subsection{CIF Score for TU\_DAT/Taekwondo Datasets}
\label{s:consistency-measure}

Table~\ref{t:main-results} shows the achieved CIF for TU\_DAT and Taekwondo dataset using X-CLIP, Video-MAE and MiniGPT4 (image-based), MiniGPT4-Video, Video-Llama and VideoMamba respectively.

It is clear that our directed FT surpasses undirected FT in all cases by a very significant margin. Note that the substantial improvement in consistency persists for two very different types of videos (road traffic vs. taekwondo), confirming that the improvement is not tied to the video characteristics. 

\begin{table}[htb]
\begin{footnotesize}
\caption{\small CIF results on TU\_DAT and Taekwondo datasets}
\setlength\tabcolsep{4.0pt} 
\begin{tabular}{|c|c|r|r|r|r|}
\hline
{\bf VLM} & {\bf Task}  & \multicolumn{2}{|c|}{{\bf Undirected}} & \multicolumn{2}{|c|}{{\bf Directed}}\\
{\bf Models} & {\bf Datasets}  & {\bf VLM$^m$} & {\bf VLM$^a$} & {\bf VLM$^m$} & {\bf VLM$^a$}\\ \hline
X-CLIP & TU\_Dat & 54.5 & 55.15 & 74.25 & 73.65\\ \hline
X-CLIP & Taekwondo & 48.71 & 48.15 & 69.85 & 70.05\\ \hline
VideoMAE & TU\_Dat & 52.04 & 52.41 & 72.65 & 73.25\\ \hline
VideoMAE & Taekwondo & 48.5 & 49.06 & 69.8 & 69.31\\ \hline
MiniGPT4 & TU\_DAT & 59.78 & 60.41 & 75.51 & 74.35\\ \hline
MiniGPT4 & Taekwondo & 59.78 & 60.41 & 75.71 & 75.41\\ \hline
MiniGPT4-Video & TU\_Dat & 71.45 & 71.8 & 86.35 & 85.125 \\ \hline
MiniGPT4-Video & Taekwondo & 68.40 & 68.1 & 83.10 & 83.08 \\ \hline
Video-Llama & TU\_Dat & 72.16 & 72.41 & 86.85 & 87.32 \\ \hline
Video-Llama & Taekwondo & 42.57 & 42.05 & 84.60 & 85.08 \\ \hline
VideoMamba & TU\_DAT & 61.95 & 61.41 & 80.85 & 80.4 \\ \hline
VideoMamba & Taekwndo & 61.60 & 61.75 & 76.52 & 75.45\\ \hline
\end{tabular}
\label{t:main-results}
\end{footnotesize}
\vspace{-12pt}
\end{table}

We also examine the impact of avoiding the use of VLM$^a$. This is possible because of the simplicity of VLM$^a$ tasks in these two datasets. Table~\ref{t:without} shows the {\em accuracy} of VLM$^m$ on test data with and without using VLM$^a$. In the latter case, the consistency check is done between the proxy activities tied to VLM$^m$ and those tied to VLM$^a$. We used MiniGPT4-Video as the main VLM in this case. The accuracy is now lower, but still substantially higher than purely undirected FT.

\subsection{CIF Results on Kinetics Dataset}
\label{s:kinetics-results}

\begin{wraptable}[6]{R}{0.45\linewidth}
\vspace{-12pt}
\footnotesize
\caption{\small FT accuracy with and w/o VLM$^a$}\label{t:without}
\vspace{-6pt}
\begin{tabular}{|l|c|c|}
\hline
{\bf Case} & {\bf TU\_DAT} & {\bf Taek.} \\ \hline
Undirected   & 75.4   & 73.41\\ \hline
With VLM$^a$ & 83.3  & 82.81 \\ \hline
W/o VLM$^a$  & 81.0   & 80.65 \\ \hline
\end{tabular}
\end{wraptable}
Unlike our first two datasets, the Kinetics dataset includes many, rather complex activities. This also generally means a more complex description for VLM$^a$, since a complex activity is best described in terms of its details, as shown in Table~\ref{t:kinetics-classes}. For example, it would be rather difficult to have a 2-3 word alternate description of "Mowing Lawn". Our method still achieves approximately the same level of accuracy for directed FT as for the simpler case of first two datasets. 

\begin{wraptable}[6]{R}{0.55\linewidth}
\vspace{-6pt}
\footnotesize{
\caption{\small CIF Results on Kinetics Dataset}\label{t:kinetics}
\vspace{-6pt}
\begin{tabular}{|l|l|l|}
\hline
{\bf VLMs} & {\bf Undirected} & {\bf Directed} \\ \hline
MiniGPT4  & 74.25\%    & 83.7 \%       \\ \hline
MiniGPT4-V  & 78.5\%    & 87.75\%        \\ \hline
Video-Mamba & 72.15\%  &  84.5\%       \\ \hline
\end{tabular}}
\end{wraptable}

Table~\ref{t:kinetics} compares the CIF measure for MiniGPT4, MiniGPT4-Video, and Videomamba. (Others are similar and not reported.)  Here we use a batch size of 50. Note that we are running directed and undirected FT for an equal amount of total time. Consequently, a significant portion of the total time is consumed in the fine-tuning of VLM$^a$. Despite this, the CIF of the directed approach is higher by about ten percentage points. This robust validation of our earlier claim that the use of VLM$^a$ can be easily justified in large, complex datasets should reassure about the soundness of our conclusions.

\subsection{Analysis of Fine-Tuning Time}

\begin{wrapfigure}[11]{R}{0.6\linewidth}
\vspace{-18pt}
\center
\includegraphics[width=\linewidth]{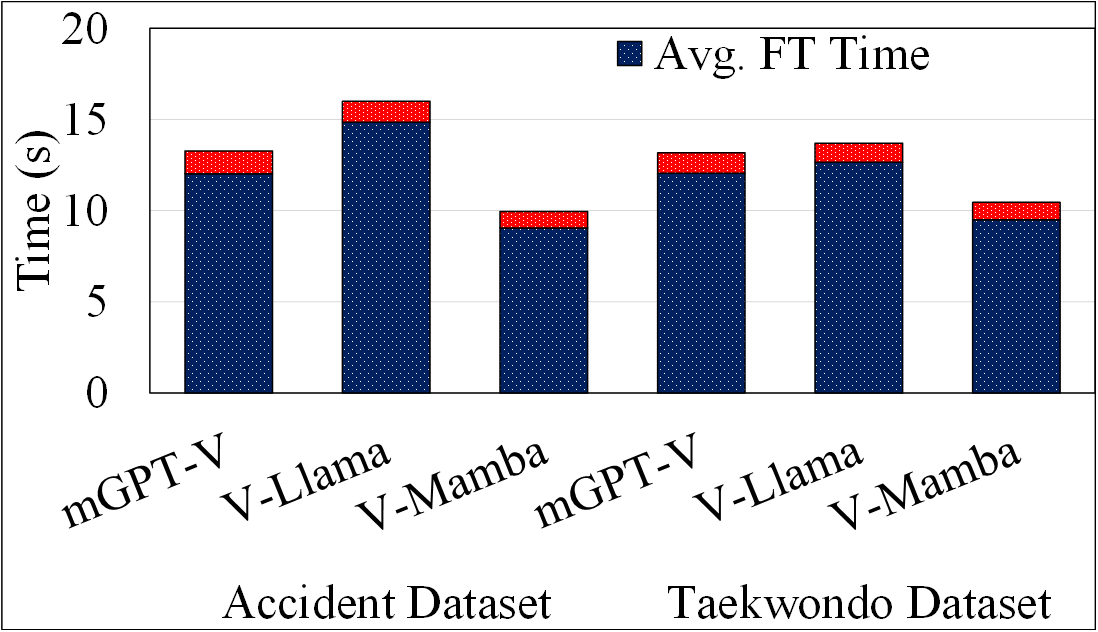}
\caption{Per-epoch prep \& FT Time for Undirected cases}
\label{f:time-total-all-u}
\end{wrapfigure}

In this section, we compare the time required for both directed and undirected FT approaches for all three datasets under consideration. As stated earlier, we use four iterations of fine-tuning for both directed and undirected cases, each using 20 videos. The time spent on each iteration consists of two parts, reported as average per epoch, over 500 epochs: (a) actual FT time and (b) preparation time. For the undirected case, prep-time randomly retrieves 20 videos from the disk. For directed cases, prep-time {\em also} includes the overhead of running YOLO, querying VLM$^m$ and VLM$^a$, generating assertions, and using them for consistency checking. 

\begin{wrapfigure}[10]{R}{0.6\linewidth}
\vspace{-6pt}
\center
\includegraphics[width=\linewidth]{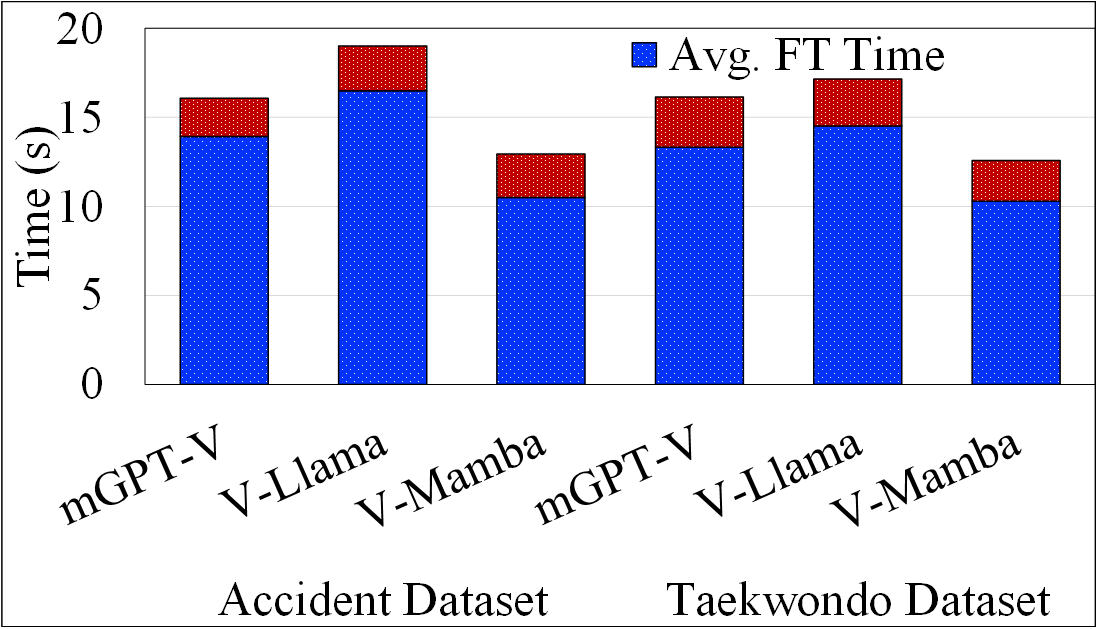}
\caption{Per-epoch prep \& FT Time for  Directed cases}
\label{f:time-total-all-d}
\end{wrapfigure} 

Fig.~\ref{f:time-total-all-u} shows the average per-epoch FT time and FT prep-time for the undirected case. Fig.~\ref{f:time-total-all-d} shows the same for the directed case. In both cases, the FT time includes both VLMs. As expected, the FT time is almost identical in both cases, and is in the $\sim$10-12 sec range (or 5-6 secs per VLM). The prep time remains small even for the directed case and not much larger than for undirected case; this is because much of it concerns retrieval and loading of videos from the disk.

\subsection{Inference and Justification Time}

\begin{wrapfigure}[13]{R}{0.6\linewidth}
 \vspace{-12pt}
\includegraphics[width=1.0\linewidth]{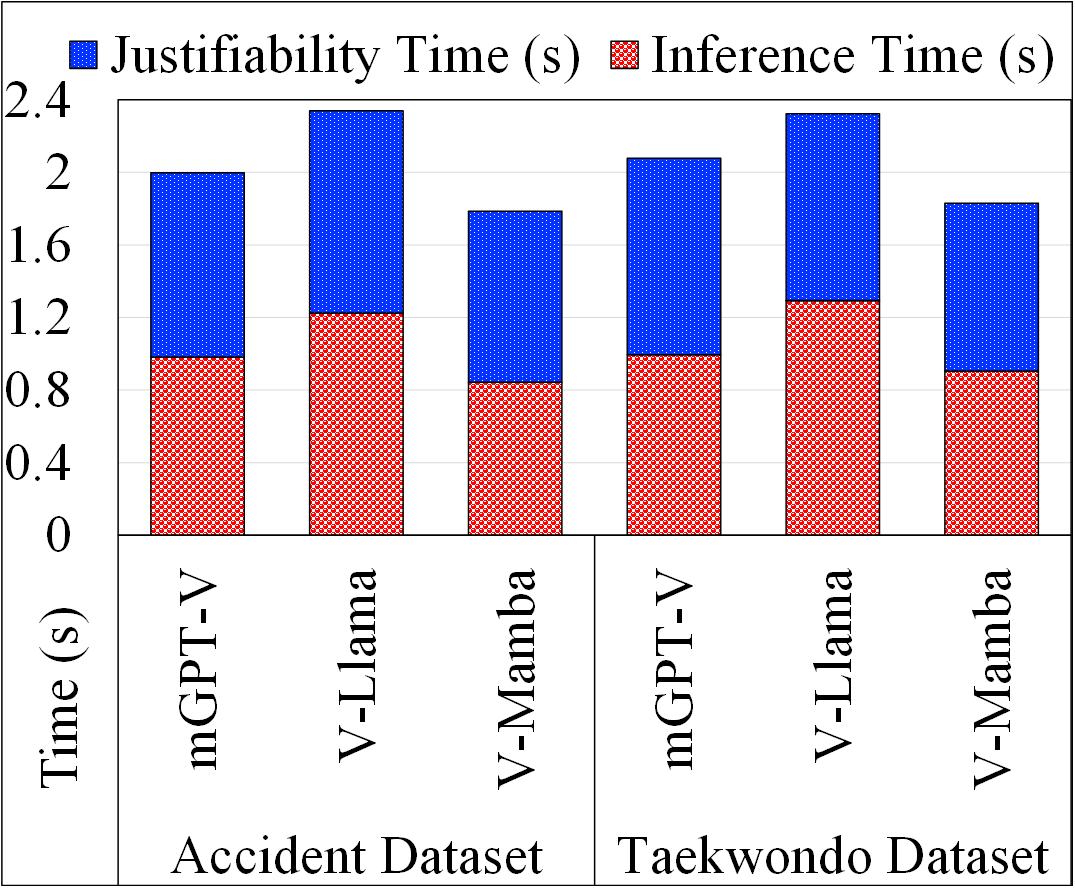}
\vspace*{-12pt}
\caption{\footnotesize{Justification/Inference Time}} \label{f:just-time}
\end{wrapfigure}

As stated earlier, we can easily turn our FT mechanism into a justifiability backed inferencing mechanism. For this, we essentially take out the FT part in Fig.~\ref{f:FT-illus}, thereby breaking the loop. In this case, each inference will also be accompanied by the result of consistency check. That is, if all consistency tests pass, we have extra backing (or justification) of the output; however, if any of them fails, we have an  indication that the output cannot be relieved upon. 

Fig.~\ref{f:just-time} displays the inference and justifiability time for MiniGPT4-Video, Video-Llama, and Video-Mamba on both datasets. Note that the justifiability time differs from the FT prep time reported above since we no longer have to pay the significant overhead of retrieving videos from disk for fine-tuning. It is seen that the justifiability time is about the same as the inference time. This should be reasonable for critical applications where inaccuracy can have serious consequences. For other applications, we may run justifiability less often (e.g., periodically or only when VLM$^m$ predicts a class that is known to have reliability issues). 

It is important to emphasize that our objective is to detect target activities for potential interventions, rather than to provide a frame-by-frame narration of events. The demonstrated inference time of approximately 2 seconds is therefore quite efficient, {\em and uses only one GPU}. While the NVIDIA A6000 GPUs used in our experiments are relatively high-end and primarily required for fine-tuning, the critical factor in deployment scenarios is inference and the models can run lower end hardware as well with corresponding increase in inference times. 

However, we believe that in a rapidly advancing field — both in terms of LLMs and AI processing hardware — it is not useful to focus on a single snapshot of the inference times obtained in this work. For example, many of our results were obtained using MiniGPT4-Video with a 7B-parameter model. More recently, Alibaba Cloud has released the Qwen2.5-VL model, also with 7B parameters. We ran several cases using Qwen2.5-VL for comparison and found that the average runtime of 1.9 seconds on MiniGPT4-Video shrinks to only 0.9 seconds on Qwen2.5-VL. Thus, the desired  goal of achieving an average inference time below 1 second is already attainable on our existing machines.

On the hardware side, the ongoing race to integrate and augment leading-edge desktop and workstation CPUs with NPUs is already making them suitable for 7B and even larger models at modest power consumption. Consider, for example, Intel’s upcoming Panther Lake workstation processor (volume production in Q1 2026), built on Intel’s 1.8 nm process. It delivers up to 180 Tera-OPS (TOPS) of ML performance — a 50\% increase over the earlier Lunar Lake processor (built on TSMC’s 3.0 nm process) — while maintaining the same 45 W thermal design power (TDP). For comparison, our NVIDIA A6000’s TOPS rating is 310 (less than 2× higher), but its TDP is 300 W (about 7× higher). Thus, Qwen2.5-VL should achieve an inference time of approximately 1.55 seconds on Panther Lake at 45 W on a laptop-to-workstation-class machine.

We would also like to point out that while there is a race toward running stripped-down LLM models on mobile and low-end IoT devices, this is not necessarily relevant for critical applications requiring very low response times. In such cases, VLMs will likely run on edge controllers rather than on individual IoT devices. Stripped-down models also tend to suffer from accuracy and hallucination issues, making them unsuitable for critical applications. Similarly, running parts of the inference process in the cloud, rather than on-premises, introduces access latency and network uncertainties — again undesirable for time-sensitive, safety-critical use cases.

\section{Conclusions}
\label{s:conc}

In this paper, we propose a novel consistency-driven fine-tuning (FT) approach for VLMs, which combines traditional computer vision (TCV) to recognize details with explicit logical reasoning to improve VLMs' performance. The mechanism can substantially reduce the labeled data needs of FT and achieve considerably higher accuracy than a mechanism that does not choose the input intelligently. It also provides a justification mechanism that can continue to be used at inference time and a vital sanity check mechanism for situational awareness applications. The proposed mechanism is quite general in that we can decide the level of complexity that we wish to introduce in consistency checking. 

The key downside of the mechanism is the need for running another VLM which increases the resource needs especially for justification during inferencing. Another limitation is that if the TCV techniques are used to go beyond the basic object/pose detection and movement tracking, some specialized training data and training will be required.


\bibliographystyle{unsrt}
\bibliography{Krishna, IoT_allrefs, AI_ML, VLM_LLM}

\end{document}